\pdfoutput=1

\documentclass[11pt]{article}

\usepackage[final]{acl}

\usepackage{times}
\usepackage{latexsym}

\usepackage[T1]{fontenc}

\usepackage[utf8]{inputenc}

\usepackage{microtype}

\usepackage{inconsolata}

\usepackage{graphicx}

\usepackage{algorithm}
\usepackage{algorithmicx}
\usepackage{algpseudocode} 

\usepackage{enumitem} 

\usepackage{booktabs} 
\usepackage{arydshln}

\usepackage{amsmath}
\usepackage{amssymb}

\usepackage{bbding} 

\newcommand\tf[1]{\textbf{#1}}
\newcommand{\authcomma}{\textmd{,}\hskip0.5em}
\newcommand{\authspace}{\textcolor{white}{\authcomma}}

\makeatletter
\def\adl@drawiv#1#2#3{%
	\hskip.5\tabcolsep
	\xleaders#3{#2.5\@tempdimb #1{1}#2.5\@tempdimb}%
	#2\z@ plus1fil minus1fil\relax
	\hskip.5\tabcolsep}
\newcommand{\cdashlinelr}[1]{%
	\noalign{\vskip\aboverulesep
		\global\let\@dashdrawstore\adl@draw
		\global\let\adl@draw\adl@drawiv}
	\cdashline{#1}
	\noalign{\global\let\adl@draw\@dashdrawstore
		\vskip\belowrulesep}}

\title{Self-Training with Direct Preference Optimization\\
Improves Chain-of-Thought Reasoning}

\author{Tianduo Wang$^{\dag}$ \authspace Shichen Li$^{\ddag}$ \authspace  Wei Lu$^{\dag}$\\
  $^\dag$StatNLP Research Group, Singapore University of Technology and Design \\
  $^\ddag$Soochow University \\
  \texttt{\{tianduo\_wang,luwei\}@sutd.edu.sg} \authspace \texttt{scli\_21@outlook.com}\\
  \url{https://github.com/tianduowang/dpo-st}}

\begin{document}
\maketitle
\begin{abstract}
	Effective training of language models (LMs) for mathematical reasoning tasks
            demands high-quality supervised fine-tuning data.
	Besides obtaining annotations from human experts, a common alternative is sampling from larger and more powerful LMs.
	%
        However, this knowledge distillation approach can be costly and unstable, 
        particularly when relying on closed-source, proprietary LMs like GPT-4~\cite{openai2023GPT4}, 
        whose behaviors are often unpredictable.
	In this work, 
		we demonstrate that the reasoning abilities of small-scale LMs can be enhanced through self-training, 
		a process where models learn from their own outputs.
	We also show that the conventional self-training can be further augmented by a preference learning algorithm called
    	Direct Preference Optimization (DPO)~\cite{rafailov2023dpo}.
	By integrating DPO into self-training, 
		we leverage preference data to guide LMs towards more accurate and diverse chain-of-thought reasoning.
	We evaluate our method across various mathematical reasoning tasks using different base models.
	Our experiments show that this approach not only improves LMs' reasoning performance 
		but also offers a more cost-effective and scalable solution compared to relying on large proprietary LMs.
\end{abstract}

\section{Introduction}

  Making language models (LMs) perform mathematical reasoning is a valuable, 
  	yet challenging research objective~\cite{hendrycks2021measuring,cobbe2021gsm8k}.
  Recent efforts have focused on enhancing large-scale LMs' reasoning abilities through various methods,
  	including chain-of-thought prompting~\cite{wei2022cot,kojima2022large}, 
  	continual pretraining~\cite{azerbayev2023llemma},
  	and adding external verifiersq~\cite{li2023making}.
  However, the research question of how to enhance the reasoning capabilities of smaller-sized LMs 
  	remains relatively under-explored.

\begin{figure}[t]
    \captionsetup{type=figure}
    \centering
    \includegraphics[width=0.95\linewidth]{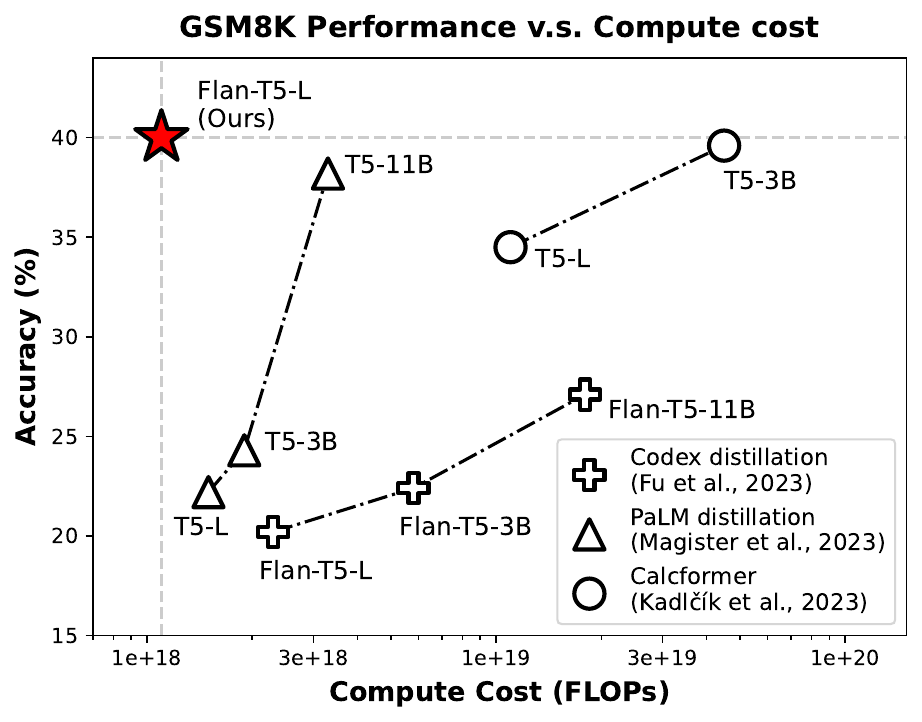}
    \caption{
        Our approach demonstrates superior performance on the GSM8K benchmark 
			while minimizing the required compute cost, including both training and inference.
        Compute cost calculations are based on the methodology outlined by~\citet{yuan2023scaling}.\protect\footnotemark
    }
    \label{fig:intro}
    \vspace{-4pt}
\end{figure}

  Recent studies~\cite{fu2023special,magister2023teaching,li2023query}
  	demonstrate that the reasoning capabilities of smaller LMs can be significantly enhanced through
  	learning from the outputs of larger and more advanced LMs,
  	such as Codex~\cite{chen2021codex}, PaLM~\cite{chowdhery2022palm}, and GPT-4~\cite{openai2023GPT4}.
  While this method is straightforward to implement, the associated costs can be substantial.
  The computational demand, measured in {\em floating-point operations} (FLOPs), increases considerably when using large LMs.
  Additionally, the reliance on proprietary large LMs for data annotation not only incurs high economic costs 
  but also raises concerns regarding the sustainability and scalability of such practices. 
  For instance, \citet{ho2023large} highlighted that while employing large LMs as annotators can largely enhance the performance of smaller LMs, 
  	it introduces a clear trade-off between economic costs and performance gains. 
  
  \footnotetext{
	All methods presented here are integrated with an external calculator except for the Codex distillation by~\citet{fu2023special}.
	}

  Another line of research focuses on exploring enhancements through 
  	self-improvement methods~\cite{zelikman2022star,gulcehre2023reinforced,singh2023beyond}.
  These methods diverge from using outputs from larger models, 
  	instead encouraging LMs to learn from their own generated data.
  The effectiveness of these techniques is evident, yet  
  	their success largely depends upon the inherent capabilities of the base models.
  For example, \citet{zelikman2022star} initiated self-improvement by few-shot prompting GPT-J~\cite{gpt-j},
	a relatively large LM which has 6 billion parameters, to generate rationales --
	an emergent ability typically reserved for large models~\cite{wei2022emergent}.
  However, the extent to which small-scale LMs can gain from self-improvement remains uncertain.

In this work,
	we introduce a novel enhancement to the conventional self-training framework by incorporating
	Direct Preference Optimization (DPO)~\cite{rafailov2023dpo}.
This integration specifically targets performance objectives within chain-of-thought reasoning, 
	with a particular focus on mathematical reasoning. 
The clear-cut nature of mathematical solutions enables straightforward validation of a model's outputs, 
	facilitating the creation of a preference dataset for DPO.
Our empirical results indicate that this method notably enhances the reasoning capabilities of LMs 
	while also reducing computational overhead.
We visualize the relationship between the GSM8K~\cite{cobbe2021gsm8k} performance
  	and computational cost across various specialized models in Figure~\ref{fig:intro}.
It can be observed that our method not only achieves strong performance,
  	but also reduces computational demands by effectively utilizing self-generated data for learning.
Overall, the main contribution of this work can be summarized as follows:
\begin{itemize}[topsep=5pt, partopsep=0pt, leftmargin=15pt, parsep=0pt, itemsep=10pt]
\item We propose a novel extension to the classic self-training framework by integrating Direct Preference Optimization, 
  demonstrating its effectiveness across various math reasoning tasks.
\item Our method significantly enhances the reasoning abilities of language models
    while requiring minimal computational resources, optimizing both performance and efficiency.
\item We present an efficient method for integrating LMs with external tools, 
  which significantly boosts downstream task performance without notably compromising inference speed.
\end{itemize}

\section{Background}\label{sec:prelim}

	\paragraph{Math word problem solving}
	The math word problem solving task~\cite{cobbe2021gsm8k} can be formulated as a sequence-to-sequence task
		where the input $x$ is a question asking for an unknown value and the output $y$ 
		is a rationale  that leads to the answer $a$.
	Normally, the answers can be extracted from the rationales via some rule-based methods, such as regular expressions.
	A generated rationale $\hat{y}$ is regarded as correct if the extracted answer $\hat{a}$ matches the gold answer $a$.
	Formally, the labeled dataset for a math word problem solving task with $l$ instances can be represented as:
	\begin{equation}
		\mathcal{L} = \{(x^i, y^i, a^i)\}_{i=1}^l\text{.}
	\end{equation}
\noindent
 A common way for specializing a LM $f_{\theta}$ towards math reasoning with the labeled dataset $\mathcal{L}$
		is {\em supervised fine-tuning} (SFT).
	It optimizes $f_{\theta}$ by minimizing the negative log likelihood loss $\mathcal{L}_{\text{SFT}}(\theta)$:
	\begin{equation}
		\mathop{\mathbb{E}}_{(x, y) \sim \mathcal{L}} -\Big[ \sum_{t=1}^T \log f_{\theta}(y_t | x, y_{1:t-1})\Big]\text{,}
	\end{equation}
	where $T$ is the length of the rationale $y$ and we use $y_t$ to represent the $t$-th token in $y$.

  \begin{algorithm}[!t]
    \centering
    \caption{Self-training for CoT reasoning tasks}
    \label{alg:st}
    \begin{flushleft}
    \hspace*{\algorithmicindent} \textbf{Input:} pre-trained language model $f_{\theta}$ \\ 
    \hspace*{\algorithmicindent} \textcolor{white}{\textbf{Input:}} labeled dataset~$\mathcal{L}=\{(x^i, y^i, a^i)\}_{i=1}^l$
    \hspace*{\algorithmicindent} \textcolor{white}{\textbf{Input:}} unlabeled dataset~$\mathcal{U}=\{(x^i, a^i)\}_{i=1}^u$
    \hspace*{\algorithmicindent} \textbf{Output:} fine-tuned model $f_{\theta'}$ \\ 
    \end{flushleft}
    \begin{algorithmic}[1]
    \State Fine-tune $f_{\theta}$ on $\mathcal{L}$ to get $f_{\theta'}$
    \Repeat
   		\State Build pseudo-labeled dataset $\mathcal{S}$:
   		\Statex~~~~~~~~~~$\mathcal{S} = \{ (x^i, \hat{y}^i, \hat{a}^i)\}_{i=1}^s$
   		\Statex~~~~~~~~~~where~~$x^i \sim \mathcal{U}$~~and~~$\hat{y}^i, \hat{a}^i \sim f_{\theta'}(\cdot|x^i)$
        \State Select $\mathcal{S}^{\alpha} \subset \mathcal{S}$~~when~~$\hat{a}^i=a^i$
        \State Update $\mathcal{L} \leftarrow \mathcal{S}^{\alpha} \cup \mathcal{L}$
        \State Train $f_{\theta}$ on $\mathcal{L}$  to get a new $f_{\theta'}$
    \Until{convergence or max iteration is reached}
  \end{algorithmic}
\end{algorithm}

  \paragraph{Self-training}
  Self-training is one of the earliest approaches in semi-supervised learning~\cite{Scudder1965ProbabilityOE, Fralick1967LearningTR} 
  	that has risen in popularity recently~\cite{He2019RevisitingSF,Amini2022SelfTrainingAS}.
  This method first regards a base model trained with a labeled dataset $\mathcal{L}$ as teacher,
  	and uses it to build a pseudo-labeled dataset $\mathcal{S}$ by annotating an unlabeled dataset $\mathcal{U}$.
  Then, a student model is trained on a combination of $\mathcal{L}$ and $\mathcal{S}$
  	that are expected to outperform the teacher model.
  Such a framework has been shown effective across a wide range of natural language processing tasks, 
  	including natural language understanding~\cite{vu2021strata} and generation~\cite{He2019RevisitingSF}.
  A formal description of a self-training algorithm for chain-of-thought (CoT) reasoning tasks is provided in Algorithm~\ref{alg:st}.

  Previous studies have demonstrated that the quality of the pseudo-labels largely
  	impacts the overall performance of the self-training algorithm~\cite{He2019RevisitingSF,Amini2022SelfTrainingAS}.
  For example,
  \citet{gulcehre2023reinforced} proposed to select high-quality pseudo-labels with a learned reward function.
  \citet{zelikman2022star} filtered the generated rationales to include only the ones that lead to correct answers.
  Although many methods are proposed to select pseudo-labels,
  	few works discuss how to improve the fine-tuned model $f_{\theta'}$ so that more high-quality pseudo-labels can be generated.
  %
  %
  In this paper,
  	we present a method to enhance $f_{\theta'}$ in each iteration so that higher-quality pseudo-labeled data can be generated.

  \paragraph{Direct Preference Optimization}
  The Reinforcement Learning from Human Feedback (RLHF) methods 
  	align LMs with human preference~\cite{Ouyang2022TrainingLM,Bai2022TrainingAH}.
  The standard pipeline of RLHF requires to first train a reward model from human preference data.
  Then, the reward model is used to fine-tune language models via reinforcement learning objective, 
  	e.g., Proximal Policy Optimization~\cite{schulman2017ppo}.
  A recent study propose Direct Preference Optimization (DPO)~\cite{rafailov2023dpo}
  	to avoid explicitly training a reward model
	so that language models can be directly tuned with human preference data.

  The DPO pipeline can be described as follows.
  First, given some prompt $x$, we sample several completions from the reference model $\pi_{\text{ref}}$
  	(normally it is the model after supervised fine-tuning):
  \begin{equation}
  	y_1, y_2 \sim 	\pi_{\text{ref}}(\cdot\ |\ x) \text{.}
  \end{equation}
  \noindent
  Next, construct the DPO dataset $\mathcal{D}$ from the completions based on the human preference:
  \begin{equation}
  	\mathcal{D} = \{(\ x^i,\ y^{i}_w,\ y^{i}_l\ )\}_{i=1}^{N} \text{,}
  \end{equation}
  where $y_w^{i}$ and $y_l^{i}$ represent the winning and losing completions respectively.
  Then, we optimize the language model $\pi_{\theta}$ to minimize $\mathcal{L}_{\text{DPO}}(\pi_{\theta};\pi_{\text{ref}})$
  	which can be defined as follows:
  \begin{equation}\label{eq:dpo}
    \mathop{\mathbb{E}}_{(x, y_w, y_l) \sim \mathcal{D}}\bigg[ -\log \sigma \Big( r(y_w | x) - r(y_l | x) \Big) \bigg] \text{,}
  \end{equation}
  where $r(\cdot | x) = \beta \log \frac{\pi_{\theta} (\cdot | x)}{\pi_{\text{ref}} (\cdot | x)} $
  	and $\beta$ is a coefficient that controls  $\pi_\theta$'s deviation from  $\pi_{\text{ref}}$.

\begin{figure*}[t]
    \captionsetup{type=figure}
    \centering
    \includegraphics[width=.86\linewidth]{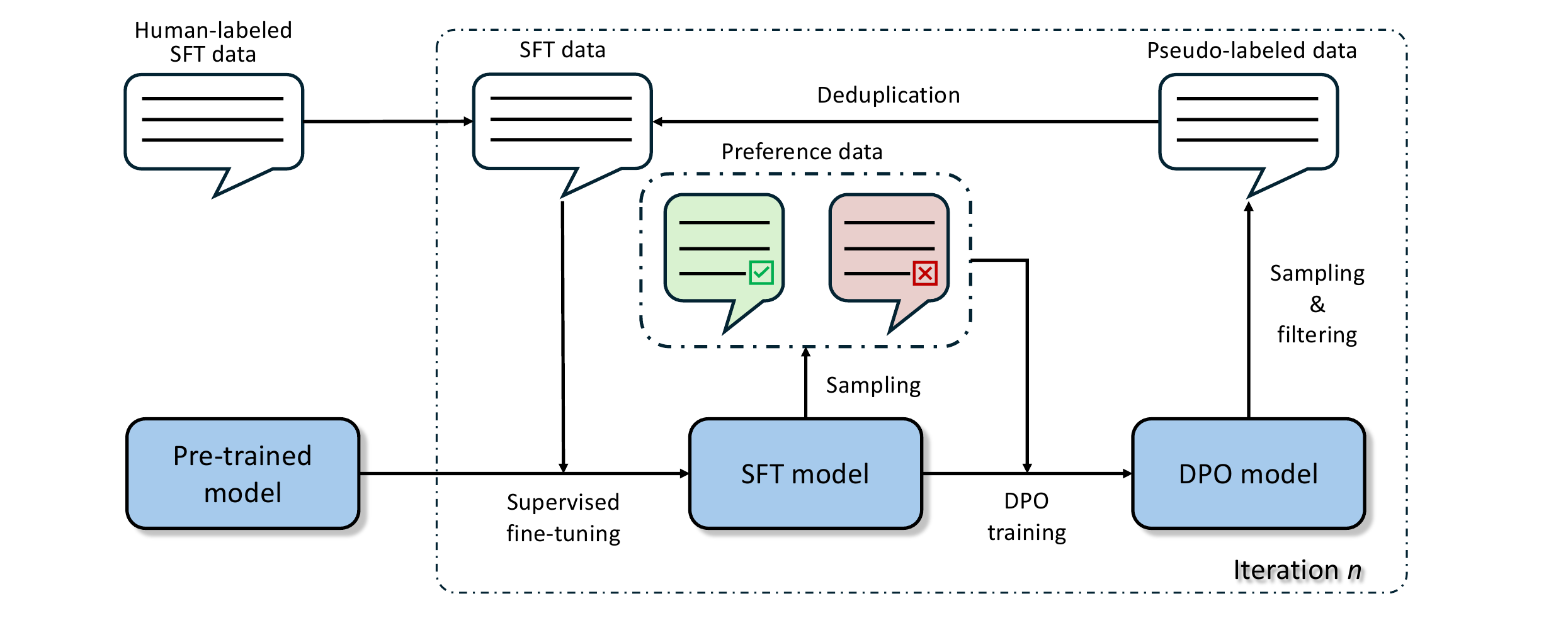}
    \caption{
        \label{fig:method} 
        An illustration of the \tf{DPO-augmented Self-Training} framework.
        Traditional self-training method
        	uses the SFT model to generate the pseudo-labels for subsequent iterations.
        In contrast,
			our method enhances the SFT model with Direct Preference Optimization (DPO), 
			using the optimized DPO model to produce the pseudo-labels.
    }
\end{figure*}

\section{Method}
	In this section,
		we first describe the proposed approach.
	Then, we demonstrate how we integrate an external calculator into the 
		model's decoding process which significantly improves LMs' performance on the downstream tasks.

	\begin{algorithm}[t]
    \centering
    \caption{DPO-augmented self-training}
    \label{alg:dpo-st}
    \begin{flushleft}
    \hspace*{\algorithmicindent} \textbf{Input:} pre-trained language model $f_{\theta}$ \\ 
    \hspace*{\algorithmicindent} \textcolor{white}{\textbf{Input:}} labeled dataset~$\mathcal{L}=\{(x^i, y^i, a^i)\}_{i=1}^l$
    \hspace*{\algorithmicindent} \textcolor{white}{\textbf{Input:}} unlabeled dataset~$\mathcal{U}=\{(x^i, a^i)\}_{i=1}^u$
    \hspace*{\algorithmicindent} \textbf{Output:} fine-tuned model $f_{\theta'}$ \\ 
    \end{flushleft}
    \begin{algorithmic}[1]
    \Statex\textcolor{lightgray}{\# Warm-up stage}
    \State Fine-tune $f_{\theta}$ on $\mathcal{L}$ to get $f_{\theta'}$
    \Repeat
    	\Statex~~~~~~\textcolor{lightgray}{\# DPO step}
    	\State Generate DPO dataset $\mathcal{D}$:
    	\Statex~~~~~~~~~~$\mathcal{D} = \{ (\ x^i,\ y_w^i,\ y_l^i\ )\}_{i=1}^N$
    	\Statex~~~~~~~~~~where~~$x^i \sim \mathcal{U}$~~and~~$y_w^i,\ y_l^i \sim f_{\theta'}(\cdot|x^i)$
    	\State Tune $f_{\theta'}$ with $\mathcal{L}_{\text{DPO}}$ on $\mathcal{D}$ to get $f_{\theta^d}$
    	\Statex~~~~~~\textcolor{lightgray}{\# SFT step}
        \State Build pseudo-labeled dataset $\mathcal{S}$:
   		\Statex~~~~~~~~~~$\mathcal{S} = \{ (x^i, \hat{y}^i, \hat{a}^i)\}_{i=1}^s$
   		\Statex~~~~~~~~~~where~~$x^i \sim \mathcal{U}$~~and~~$\hat{y}^i, \hat{a}^i \sim f_{\theta^d}(\cdot|x^i)$
        \State Select $\mathcal{S}^{\alpha} \subset \mathcal{S}$~~when~~$\hat{a}^i=a^i$
        \State Update $\mathcal{L} \leftarrow \mathcal{S}^{\alpha} \cup \mathcal{L}$
        \State Train $f_{\theta}$ on $\mathcal{L}$  to get a new $f_{\theta'}$
    \Until{convergence or max iteration is reached}
	\end{algorithmic}
	\end{algorithm} 

\subsection{DPO-augmented Self-Training}\label{sec:method}
	
	Our approach starts with a {\em warm-up stage},
		and then followed by an iterative process,
		where each iteration is composed of two sub-steps: {\em DPO step} and {\em SFT step}.
	The iterative process ends when the model performance converges or reaches the maximum iteration.
	A formal description of the proposed method is illustrated in Algorithm~\ref{alg:dpo-st}.
    An illustration of our method is presented in Figure~\ref{fig:method}.

	\paragraph{Warm-up stage}
	Like classic self-training, we start by fine-tuning the base model $f_{\theta}$ 
		to optimize $\mathcal{L}_{\text{SFT}}(\theta)$ on the labeled data $\mathcal{L}$, resulting in an updated model
		$f_{\theta'}$.
	After this stage, we assume that $f_{\theta'}$ is capable of solving certain math problems.
	Specifically, given a math question $x$, $f_{\theta'}$ will generate a rationale $\hat{y}$ with answer $\hat{a}$.

	\paragraph{Iterative step 1: DPO step}
	In this step, 
		we first sample rationales $\hat{y}$ from the fine-tuned model $f_{\theta'}$ given some questions $x$ from $\mathcal{U}$.
	For each question $x$, we generate multiple rationales to build the DPO training dataset $\mathcal{D}$.
	As mentioned, for math problem solving tasks,
		it is easy to know whether a generated rationale $\hat{y}$ can be considered as correct.
	We label rationales with correct answers as winning completions, 
		while consider rationales with incorrect answers as losing completions.
	Then, we train $f_{\theta'}$ on $\mathcal{D}$ to optimize the objective function $\mathcal{L}_{\text{DPO}}$
		and get a DPO model $f_{\theta^d}$ in the end.

	\paragraph{Iterative step 2: SFT step}
	After obtaining $f_{\theta^d}$,
		we use it to generate a new pseudo-labeled dataset $\mathcal{S}$ for the next-round supervised fine-tuning:
	\begin{equation}
			\mathcal{S} = \{ (x, \hat{y}) | x\sim\mathcal{U}, \hat{y} \sim f_{\theta^d}(\cdot | x) \}
	\end{equation}

 	After generation,
		we clean $\mathcal{S}$ by eliminating rationales with incorrect answers and removing duplicates.
	Therefore, the pseudo-labeled dataset we obtained in the end is a subset of the original one, 
		i.e., $\mathcal{S}^{\alpha} \subset \mathcal{S}$.
	The final training dataset is the combination of the original labeled dataset $\mathcal{L}$ 
		and the newly-generated pseudo-labeled dataset $\mathcal{S}^{\alpha}$.
	Notice that during this process,
		once we collect a new dataset, we train from the original base model $f_{\theta}$ instead of
		continually fine-tuning $f_{\theta'}$ to avoid overfitting, following previous practice~\cite{zelikman2022star,singh2023beyond}.

	\subsection{Batch Decoding with Calculator}\label{sec:calc}
			
		Empirical observations indicate that
			while large LMs, such as those described in~\citet{brown2020gpt3},
			demonstrate superior proficiency in basic arithmetic calculations,
			smaller LMs like Flan-T5-Large tend to struggle with similar arithmetic tasks.
		This limitation significantly affects their performance in math reasoning tasks.
		To address this,
			various studies~\cite{parisi2022talm,schick2023toolformer,kadlcik2023calcx} have explored 
			augmenting small-scale models with an external calculator to boost their arithmetic capabilities.
		However, many of these existing methods are limited to a batch size of one during decoding.
		This constraint substantially reduces the inference speed and limits their practical application.

		\begin{figure}[t]
			{
				\texttt{Q: James writes a 3-page letter to 2 different friends twice a week. How many pages does he write a year?}\\
				\texttt{A: He writes each friend}\\
				\texttt{3*2=\textcolor{blue}{<{}<3*2=6>{}>}6 pages a week.}\\
				\texttt{So he writes}\\
				\texttt{6*2=\textcolor{blue}{<{}<6*2=12>{}>}12 pages every week.}\\
				\texttt{That means he writes}\\
				\texttt{12*52=\textcolor{blue}{<{}<12*52=624>{}>}624 pages a year.}\\
				\texttt{\#\#\#\# 624}
				\caption{
				\label{fig:example}
					An example from the GSM8K dataset.
					The calculation annotations are highlighted in \textcolor{blue}{blue}.
					All calculation steps are wrapped within special tokens \textcolor{blue}{\texttt{<{}<...>{}>}}.
					During decoding, the calculator will be triggered when such patterns exist
						and the model's output tokens will be overridden by the calculator results.
					Following \citet{cobbe2021gsm8k}, 
						the calculation is performed with the in-built python function \texttt{eval()}.
					}
			}
			\end{figure}

		To address this challenge, we propose a simple yet efficient method that 
			allows for using larger batch sizes during inference with an external calculator.
		Our approach leverages the calculator annotations provided in the original GSM8K dataset~\cite{cobbe2021gsm8k}.
		Figure~\ref{fig:example} demonstrates an example of this annotation and
			describes how such annotations can be used during decoding.
		For optimal utilization of these annotations, we build our models with the Transformers library~\cite{wolf2020hf}.
		During inference, we employ a customized 
			\texttt{LogitsProcessor}\footnote{
				\url{https://huggingface.co/docs/transformers/internal/generation_utils\#logitsprocessor}
				}--available in the Transformers documentation--
			to adjust the model's generation process.
		This \texttt{LogitsProcessor} acts as an interface, 
			allowing modifications to the outputs of the model during generation and 
			thereby enabling efficient management of larger batch sizes.

\begin{figure}[t]
    \captionsetup{type=figure}
    \centering
    \includegraphics[width=.93\linewidth]{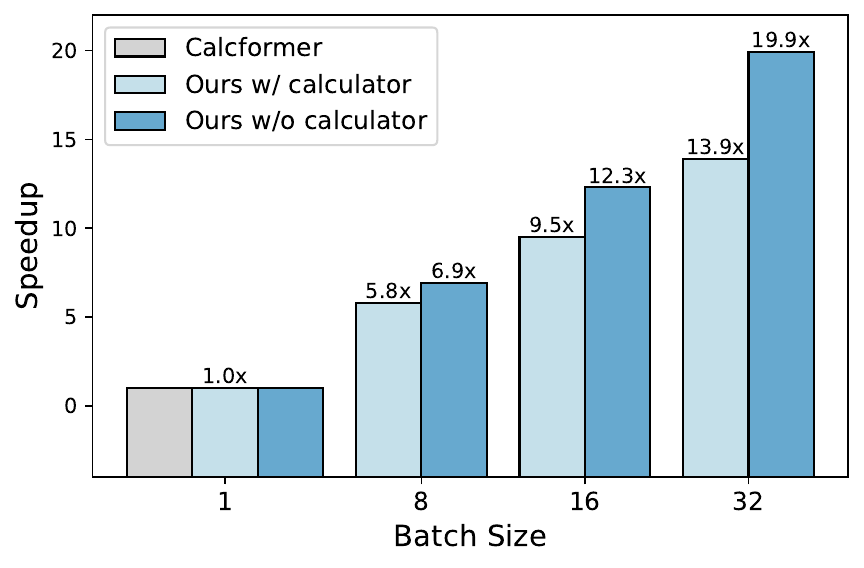}
    \caption{
        \label{fig:speed} 
        Inference speed comparison between our methods (w/ and w/o calculator) and Calcformer~\cite{kadlcik2023calcx}
        	with varying batch sizes.
        The results are measured on a single NVIDIA A40 GPU.
    }
\end{figure}

		To demonstrate the efficiency of the proposed solution,
			we compare the inference speed of our methods (w/ and w/o calculator) based on Flan-T5-Large against an open-source tool-using method, 
			Calcformer~\cite{kadlcik2023calcx} based on T5-Large, in Figure~\ref{fig:speed}.
		We find that when the batch size equals 1,
			all three methods have a similar inference speed of around 40 tokens per second.
		However, as the inference batch size increases,
			the speedup of our methods increases significantly.

\section{Experiments}
	In this section, we first outline our experiment setup and implementation details, 
		then present our models' performance on various math reasoning tasks against competitive baselines.
	Finally, we analyze the effectiveness of our method empirically.

\subsection{Setup}\label{sec:setup}

	\paragraph{Base models}
	We employ Flan-T5 models~\cite{chung2024scaling} as our primary base models.
	Specifically, we consider two variants from the Flan-T5 family:
		Flan-T5-Base and Flan-T5-Large.
	We select Flan-T5 over the original T5 models~\cite{t5} as our backbone models
		based on the evidence from previous research~\cite{chung2024scaling,fu2023special}, 
		which demonstrates that instruction-tuned models like Flan-T5
		outperform their pre-trained counterparts in mathematical reasoning tasks.
	To broaden our analysis, we also include Llama models~\cite{touvron2023llama1,touvron2023llama,llama3} 
		as additional base models for comparison.

\begin{table}[t]
    \begin{center}
    \centering
    \resizebox{\linewidth}{!}{
        \begin{tabular}{llr}
        \toprule
           \tf{Dataset} &  
           \tf{Split} &
           \tf{\# Data}  \\
        \midrule
        
        	GSM8K~\cite{cobbe2021gsm8k}         
        				  &   Train      &       6,705\\
        	              &   Validation &        \textcolor{white}{0,}768\\
	                      &   Test       &       1,319\\
	        
	        \midrule
            MultiArith~\cite{roy2015mwp}       &   Test &       \textcolor{white}{0,}600\\
            ASDiv~\cite{miao2020asdiv}         &   Test &       2,096 \\
            SVAMP~\cite{patel2021svamp}        &   Test &       1,000 \\
        \bottomrule
        \end{tabular}
    }
    \caption{
        \label{tab:dataset_stat}
        Statistics of the datasets used in our experiments.
        The original GSM8K dataset only contains train and test split.
        We randomly select 768 training examples to construct the validation dataset in our experiments.
    }
    \end{center}
	\vspace{-5pt}
\end{table}

\begin{table*}[t]
\centering
\scalebox{0.90}{
\begin{tabular}{lcccccc}
    \toprule
    \textbf{Method}& \textbf{Base Model} & \textbf{GSM8K} & \textbf{MultiArith} & \textbf{ASDiv} & \textbf{SVAMP}\\
    \toprule

    Supervised Fine-Tuning
    & Flan-T5-Base & 18.1 & 54.2 & 26.2 & 19.5 \\
    
    Self-Training
    & Flan-T5-Base & 25.9 & 73.8 & 28.2 & \tf{24.2} \\

	DPO-aug Self-Training ({\em Ours})
    & Flan-T5-Base & \tf{27.2} & \tf{74.3} & \tf{29.2} & 22.6\\
    \midrule
    
    Supervised Fine-Tuning
    & Flan-T5-Large & 30.8 & 77.2 & 38.1 & 33.6 \\
	
	Self-Training
    & Flan-T5-Large & 35.6 & 86.2 & 42.5 & 34.8\\
	DPO-aug Self-Training ({\em Ours})
    & Flan-T5-Large & \tf{37.4} & \tf{89.0} & \tf{42.8} & \tf{36.8}\\  
    
    \bottomrule
\end{tabular}
}
\caption{
        \label{tab:main_results}
       	Overall accuracies (\%) over four math word problem solving tasks.
       	Inspired by the previous practice~\cite{fu2023special}, 
       		all the models in this table are only trained with the GSM8K training set~\cite{cobbe2021gsm8k}.
       	Hence, we report the in-distribution performance for GSM8K,
       		while reporting the out-of-distribution performance for the other three datasets,
       		i.e., MultiArith, ASDiv, and SVAMP.
    }
\end{table*}

\paragraph{Datasets}
The labeled dataset $\mathcal{L}$ used in our experiments
	comes from the training split of the GSM8K dataset.
Our unlabeled dataset $\mathcal{U}$
	is also built upon GSM8K's training data by removing its annotated rationales.
For evaluation, we consider three additional commonly used math reasoning tasks besides GSM8K: 
	MultiArith, ASDiv, and SVAMP. 
Table~\ref{tab:dataset_stat} provides the statistics information of each dataset.
Following previous practice~\cite{fu2023special},
	we fine-tune our base models exclusively on the GSM8K training data while utilizing the rest three datasets
	to evaluate our models' out-of-domain performance
	as they do not have an official in-domain training split.

\subsection{Implementation Details}

In the warm-up stage, we fine-tune the base models on the training set of GSM8K~\cite{cobbe2021gsm8k}
	with the original human-labeled annotations and obtain the initial SFT model.
For subsequent DPO steps,
	we first sample rationales from SFT models to build the preference dataset.
We sample 5 rationales per question with a temperature of 0.7.
Generated rationales $\hat{y}$ containing the correct answer are classified as winning ones $y_w$, 
while the rest are considered losing ones $y_l$.
We set $\beta=0.1$ in the DPO learning objective $\mathcal{L}_{\text{DPO}}$.
For the subsequent SFT steps,
	we generate 3 rationales per question from the DPO-tuned model $f_{\theta^d}$, also with a temperature of 0.7.
Only the correct generated rationales $\hat{y}$ will be selected to build the pseudo-labeled dataset.
For both DPO and SFT steps, we perform simple deduplication based on the Jaccard similarity scores with a threshold of 0.7.
Additional implementation details can be found in Appendix~\ref{app:impl}.

\paragraph{Baselines}
We mainly consider two baseline methods to compare with our method: Supervised Fine-Tuning (SFT) and Self-Training (ST).
The SFT baseline corresponds to the model after the warm-up stage.
The Self-Training baseline adheres to the procedure outlined in Algorithm~\ref{alg:st}.
To ensure a fair comparison between our proposed method and the ST baseline,
	we use the same set of hyperparameters for both methods at each iteration.

\subsection{Main Results}		
	
	\paragraph{Comparison with baselines}
	Table~\ref{tab:main_results} shows the performance of our method compared with the baselines
		using two base models, Flan-T5-Base and Flan-T5-Large, across four datasets.
	The results clearly show that both the ST baseline and our proposed DPO-augmented Self-Training method
		outperform the SFT baseline by a large margin,
		indicating the effectiveness of the self-training framework in general.
	Although the ST baselines make significant improvements over the SFT baselines,
	our DPO-augmented Self-Training models demonstrate enhanced performance on 
        both in-domain (GSM8K) and out-of-domain (MultiArith, ASDiv, and SVAMP) tasks.

	\begin{table*}[t]
\centering
\scalebox{0.80}{
\begin{tabular}{lrcccr}
    
    \toprule
   
	\textbf{Method}
	& \textbf{Base Model}  & \textbf{\# Annotations} & \textbf{Annotator} & \textbf{Tools} & \textbf{Acc.}\\
    
    \toprule
    
    \multicolumn{6}{l}{\it{Supervised fine-tuning}} \\
	
	CoT~\cite{shridhar2023distilling}
    & GPT-2-Large & \textcolor{white}{00}7K & Human & \XSolidBrush & 14.1\\
    
    Self-consistency~\cite{khalifa2023discriminator}
    & Flan-T5-Large & \textcolor{white}{00}7K & Human & \Checkmark  & 33.3 \\
    
    GRACE~\cite{khalifa2023discriminator}
    & Flan-T5-Large & \textcolor{white}{00}7K & Human & \Checkmark  & 36.3 \\
    
    Calcformer~\cite{kadlcik2023calcx}
    & T5-Large & \textcolor{white}{0}30K & Human & \Checkmark  & 34.2 \\
    
    \midrule
    \multicolumn{6}{l}{\it{Knowledge Distillation}} \\ 
    
    Socratic CoT~\cite{shridhar2023distilling}
    & GPT-2-Large & \textcolor{white}{00}7K & GPT-3 175B & \XSolidBrush & 21.1\\
    
    CoT from CodeX~\cite{fu2023special}
    & Flan-T5-Large & 100K & CodeX & \XSolidBrush & 20.2\\
    
    CoT from PaLM~\cite{magister2023teaching}
    & T5-Large & \textcolor{white}{00}6K & PaLM 540B & \Checkmark & 22.2 \\
    
    \midrule
    \multicolumn{6}{l}{\it{Ours}} \\ 

	DPO-aug Self-Training~({$K$=3})
    & Flan-T5-Large & \textcolor{white}{00}7K & Human & \Checkmark & 37.4 \\
    DPO-aug Self-Training~({$K$=5})
    & Flan-T5-Large & \textcolor{white}{00}7K & Human & \Checkmark & 39.1 \\
    DPO-aug Self-Training~({$K$=10})
    & Flan-T5-Large & \textcolor{white}{00}7K & Human & \Checkmark & \tf{40.0} \\

    \bottomrule
\end{tabular}
}
\caption{
        \label{tab:sota}
        Detailed comparison among existing methods with comparable model sizes on the GSM8K test set.
        The ``Annotator'' column indicates how the rationales of the training data are generated.
        In this column, ``Human'' refers to the labels from the original GSM8K dataset~\cite{cobbe2021gsm8k} 
        	that are written by human annotators.
	    The ``Tools'' column indicates whether external calculators are applied during inference.
    }
\end{table*}

\begin{figure}[t]
    \captionsetup{type=figure}
    \centering
    \includegraphics[width=.94\linewidth]{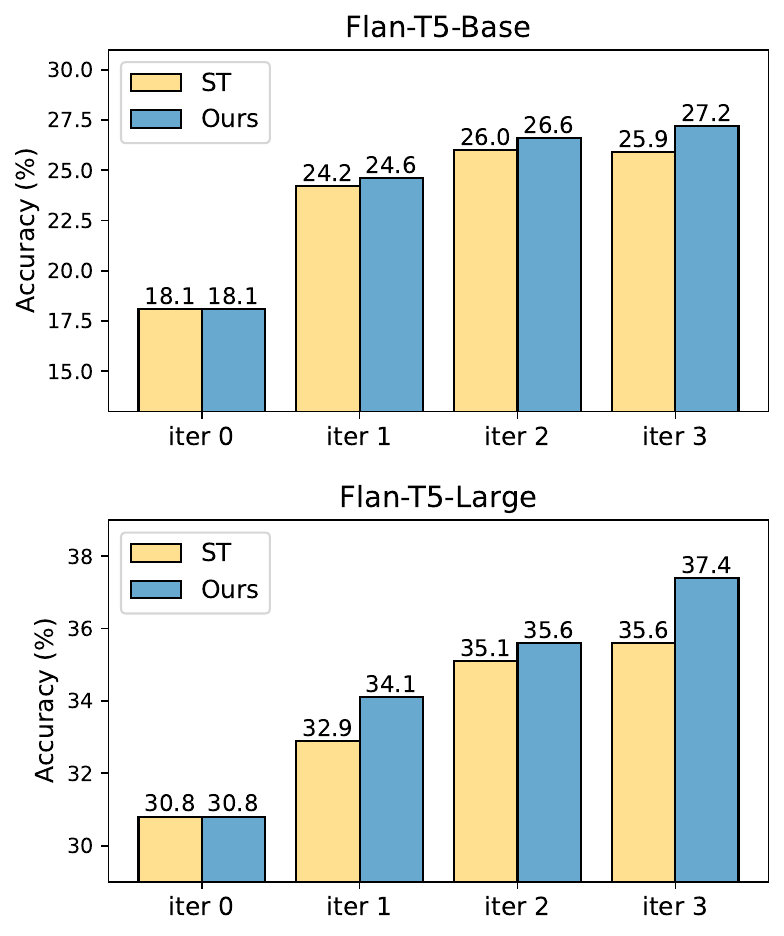}
    \caption{
        \label{fig:iter} 
        The performance of the proposed method on GSM8K over three iterations.
		For ``iter 0'', we report the performance of the SFT baselines, 
		which are obtained after the warm-up stage.
    }
	\vspace{-8pt}
\end{figure}
		
	\paragraph{Effect of iterative training}
	Figure~\ref{fig:iter} demonstrates the impact of iterative training on Flan-T5-Base and Flan-T5-Large models,
		comparing our method to the ST baseline.
	Initially, both methods start with a warm-up stage and have similar accuracies at iteration 0.
	As training progresses, our method consistently outperforms ST across iterations for both models. 
	For Flan-T5-Base, the accuracy improvement plateaus by iteration 3, suggesting convergence.
	In contrast, Flan-T5-Large shows a clear and steady improvement, 
		with our method achieving significantly higher accuracy by iteration 3. 
	This underscores the effectiveness of our iterative training process, 
		particularly in enhancing performance of larger models.

\subsection{Comparison with Existing Methods}
In this section, we compare our methods with existing approaches.
To enhance our method, 
we increase the number of sampled pseudo-labels per question to build a more diverse and robust
pseudo-label dataset.
We denote this hyperparameter as $K$ following~\citet{yuan2023scaling}.
		
Table~\ref{tab:sota} presents a detailed comparison between our method
    and exisiting methods using a simialr base model size.
The base models we considered include GPT-2-Large~\cite{gpt2},
    T5-Large~\cite{t5}, and Flan-T5-Large~\cite{chung2024scaling},
	each with approximately 770 million parameters.
As shown in Table~\ref{tab:sota},
    our approach not only outperforms other methods on the GSM8K benchmark,
    but also demonstrates remarkable label efficiency by exclusively using the annotations from the original GSM8K dataset.

In Table~\ref{tab:llama_compare}, 
	we further evaluate the effectiveness of the proposed method 
	with the Llama model family~\cite{touvron2023llama1,touvron2023llama,llama3},
    comparing it with several state-of-the-art closed-source models
    as well as similarly sized open-source models.
We observe a substantial performance gap between proprietary and open-source models.
Among the open-source models,
	those utilizing knowledge distillation generally outperform their counterparts without such enhancement.
Notably, our models using Llama-1-7b and Llama-2-7b base models 
	surpass other open-source alternatives 
	that do not employ knowledge distillation, achieving accuracies of 44.7\% and 54.7\% respectively.
Furthermore, our model employing the latest Llama-3-8b~\cite{llama3}
	matches or exceeds the performance of earlier models with knowledge distillation,
	demonstrating a significant accuracy of 68.8\%.

\begin{table}[H]
    \begin{center}
    \centering
    \resizebox{0.97\linewidth}{!}{
        \begin{tabular}{lcr}
        \toprule
           \tf{Method} &  \tf{Base Model} & \tf{Acc.}  \\
        \midrule
        	\multicolumn{3}{l}{\it{Closed-source models}}\\
        	Claude-3-Opus~\cite{claude-3}     &   -     &       95.0\\
        	Claude-2~\cite{claude-2}          &   -     &       88.0\\
        	GPT-4~\cite{openai2023GPT4}       &   -     &       92.0\\
        	Flan-PaLM-2~\cite{anil2023palm}	  &   -     &       84.7\\
        \midrule
        	\multicolumn{3}{l}{\it{Open-source models w/ knowledge distillation}}\\
	        MAmooTH~\cite{yue2023mammoth}$^\heartsuit$ & Llama-2-7b & 53.6\\
	        LEMA~\cite{an2023learning} & Llama-2-7b & 54.1\\
	        WizardMath~\cite{luo2023wizardmath} & Llama-2-7b & 54.9\\
	        MetaMath~\cite{yu2023metamath} & Llama-2-7b & 66.5\\
	        MuggleMath~\cite{li2023query} & Llama-2-7b & 68.4\\
	        ToRA~\cite{gou2023tora}$^\heartsuit$ & Llama-2-7b & 68.8\\
	        \midrule
	        \multicolumn{3}{l}{\it{Open-source models w/o knowledge distillation}} \\
	        SFT~\cite{yuan2023scaling}   &   Llama-1-7b      &       35.9\\ 
			SFT w/ Calculator$^\heartsuit$   &   Llama-1-7b      &       40.0\\
			RFT ($K$=100)~\cite{yuan2023scaling}   &   Llama-1-7b      &       41.7\\ 
			SFT~\cite{yuan2023scaling}   &   Llama-2-7b      &       41.6\\
			SFT w/ Calculator$^\heartsuit$   &   Llama-2-7b      &       45.1\\
			RFT ($K$=100)~\cite{yuan2023scaling}	 &   Llama-2-7b      &       47.5\\
			SFT w/ Calculator$^\heartsuit$   &   Llama-3-8b      &       61.0\\
        	\cdashlinelr{1-3}
        	\multicolumn{3}{l}{\it{Ours}} \\
        	DPO-ST ($K$=10)$^\heartsuit$             &   Llama-1-7b      &       44.7\\
        	DPO-ST ($K$=10)$^\heartsuit$             &   Llama-2-7b      &       54.7\\
        	DPO-ST ($K$=10)$^\heartsuit$             &   Llama-3-8b      &       \tf{68.8}\\
        \bottomrule
        \end{tabular}
    }
    \caption{
        \label{tab:llama_compare}
        Comparison with the state-of-the-art proprietary models and
			Llama-based open-source models~\cite{touvron2023llama1,touvron2023llama,llama3}.
		$^\heartsuit$: models augmented with external tools.
    }
    \end{center}
    \vspace{-5pt}
\end{table}

\subsection{Effects of the DPO Step}
\begin{figure}[t]
    \captionsetup{type=figure}
    \centering
    \includegraphics[width=0.95\linewidth]{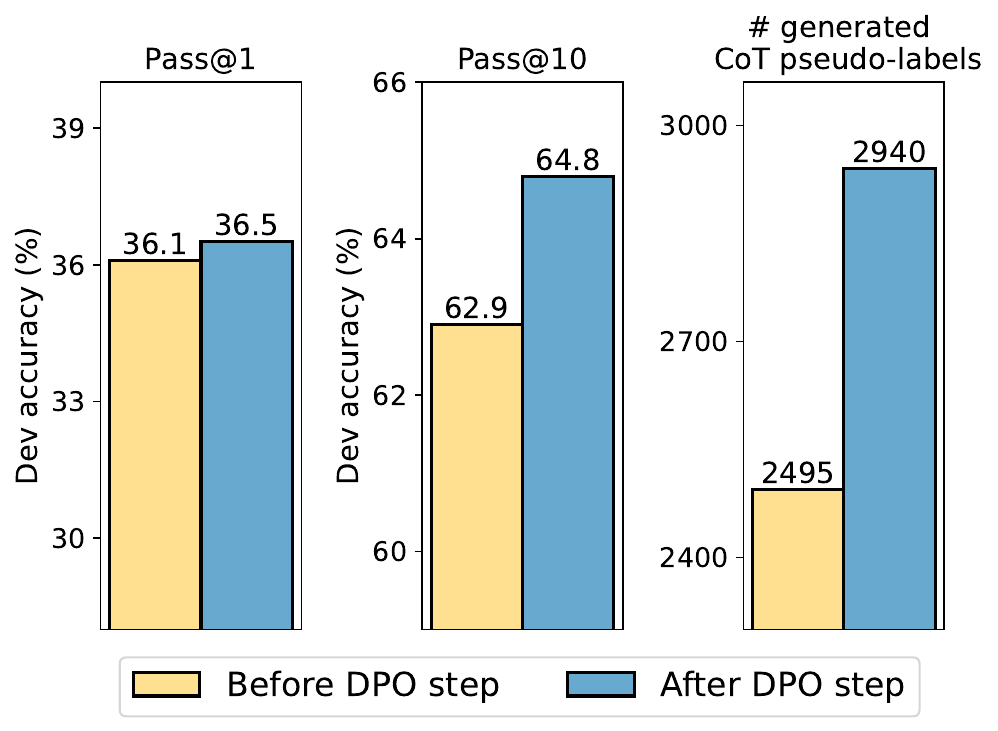}
    \caption{
        \label{fig:dpo_step} 
        Effects of the DPO step.
        \textbf{Left:}
        	we report the greedy decoding results for Pass@1.
        \textbf{Middle:} For Pass@10,
        	the solutions are sampled with temperature 0.7.
        \textbf{Right:} We count the number of generated pseudo-labels
        	after deduplication.
    }
\end{figure}

	As mentioned earlier,
		the main difference between the proposed method and
		the classic self-training is the DPO step in every iterative process.
	We now analyze how the DPO steps improve self-training.
	Figure~\ref{fig:dpo_step}
		compares the performance of models before and after the DPO step in the first iteration
		on the Pass@K metrics.
	Pass@K measures the probability that
		at least one of the $K$ generated solutions for a problem is correct,
		which serves as a gauge for both the quality and the variety of the model-generated solutions.
	The models we investigate here are fine-tuned from the Flan-T5-Large.

	As shown in Figure~\ref{fig:dpo_step},
		the DPO step yields only marginal improvements over
		the SFT model in the Pass@1 performance on the development set.
	However, the performance significantly improves
		when multiple rationales, i.e., 10 solutions per question, are sampled with temperature 0.7 (measured with the Pass@10 metric).
	This indicates that the DPO training objective makes language models inclined to generate rationales of both high quality and diversity.
	We also compare the number of generated rationales on the training set $\mathcal{L}$
		for models with and without the DPO step.
	Figure~\ref{fig:dpo_step} (right) clearly shows that
		the model after the DPO step can produce more SFT data for the next iteration.

\subsection{Effects of External Calculator}
\begin{figure}[t]
    \captionsetup{type=figure}
    \centering
    \includegraphics[width=0.93\linewidth]{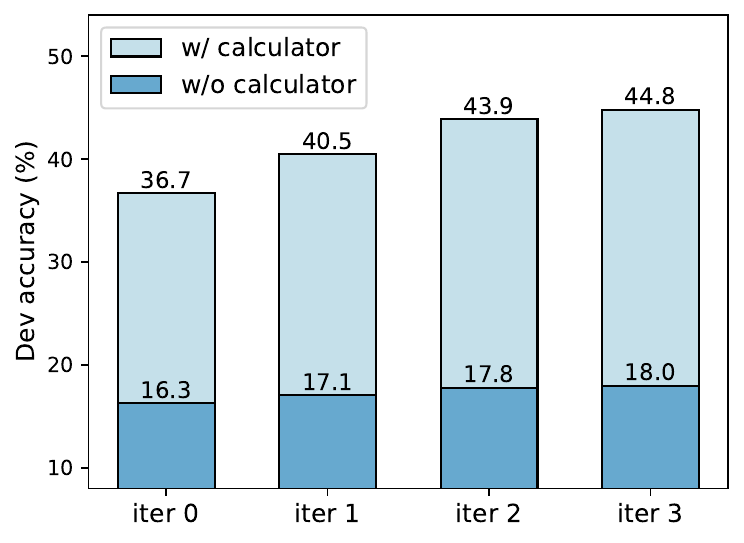}
    \vspace{+3pt}
    \caption{
        \label{fig:calculator}
		GSM8K development set accuracy of Flan-T5-Large 
		with and without the use of an external calculator during inference.
    }
\end{figure}

	Driven by the observation that
		small-scale LMs frequently make basic calculation errors,
		we develop a simple yet efficient method that 
		integrates an external calculator into the models' decoding process.
	To evaluate the impact of this integration, 
		we conduct an ablation study by omitting the calculator and present the findings in Figure~\ref{fig:calculator}.
	Our results indicate that decoding without the calculator markedly reduces accuracy across all iterations.
	We believe that this is because
		models will generate large amount of false positive pseudo-labels without calculator, 
		that is,
		the generated pseudo-labels may have correct final answers 
		but make errors in the intermediate reasoning steps.

\section{Related Work}

\paragraph{Learning from pseudo-labels}
Supervised fine-tuning (SFT)
	is prevalent technique employed to enhance the performance of 
	pre-trained language models on specific downstream tasks~\cite{Ouyang2022TrainingLM,chung2024scaling}.
However, this method heavily depends on the availability of high-quality labeled data, 
	which can be both expensive and labor-intensive to procure~\cite{brown2020gpt3}.
To address this limitation,
	various strategies have been developed to generate high-quality pseudo-labels
	using either unlabeled or synthetic data for a wide range of applications,
	including
	text classification~\cite{xie2020unsupervised},
	sentence representation learning~\cite{wang2022diffaug},
	instruction tuning~\cite{honovich2022unnatural},
	and math reasoning~\cite{wang2023msat}.
Recent advancements in this area primarily focus on two directions:
	self-training and knowledge distillation.
The key difference between these methods lies in the source of the pseudo-labels:
	self-training uses the model's own predictions on unlabeled data, 
	while knowledge distillation utilizes the insights from larger, more powerful models.

\paragraph{Self-training in language model}
Recently, we have witnessed a large number of works focusing on self-training algorithms for language models
\cite{He2019RevisitingSF,zelikman2022star,yuan2023scaling}.
Most of such methods are built upon the classic self-training framework~\cite{Scudder1965ProbabilityOE}. 
%
\citet{He2019RevisitingSF} empirically studied the effectiveness of self-training in natural language generation tasks, 
	e.g., summarization and translation.
\citet{zelikman2022star} proposed {\em self-taught reasoner} (STaR),
	which demonstrated that language models can be iteratively improved from its own generation,
	even there are no gold rationales provided.
\citet{yuan2023scaling} proposed {\em rejection sampling fine-tuning} to improve language models'
math reasoning abilities.
This method can be interpreted as only executing one iteration of the self-training algorithm.
\citet{singh2023beyond} proposed ReST$^{EM}$, a self-improving algorithm based on expectation-maximization framework.
This method demonstrates significant improvements in problem-solving tasks, e.g., math reasoning and code generation.

\paragraph{Knowledge distillation from LLMs}
Many of the recent research efforts demonstrated large language models (LLMs)
	are capable of performing math reasoning~\cite{wei2022cot,gao2022pal,openai2023GPT4,anil2023palm}.
As a result, there is growing interest in enhancing the reasoning abilities of smaller language models 
	by distilling chain-of-thought pseudo-labels from LLMs.~\cite{ho2023large,magister2023teaching,fu2023special}.
For example,
\citet{luo2023wizardmath} proposed Reinforcement Learning from Evol-Instruct Feedback
	built upon the Evol-Instruct framework~\cite{xu2023wizardlm},
	which requires ChatGPT to provide the training signals.
\citet{an2023learning} demonstrated that language models can effectively learn from the
	mistakes that can be corrected by LLMs during supervised fine-tuning.
Although these methods are shown to have promising experimental results,
	they are costly to implement as large models cost more FLOPs during inference.
Our work demonstrates that small-scale language models can effectively learn from their own generations,
	offering a more resource-efficient alternative to knowledge distillation.
Since our method is conceptually orthogonal to knowledge distillation techniques,
	an interesting avenue for future research would be to explore integrating knowledge distillation into our iterative training process 
	to further enhance model performance.

\section{Conclusion}

We present an effective and resource-efficient method called DPO-augmented Self-Training (DPO-ST), 
	which augments the original Self-Training algorithm with Direct Preference Optimization~\cite{rafailov2023dpo}.
Unlike previous studies that improve small-scale language models' reasoning abilities 
    by distilling a larger and more powerful model,
    we argue that small models that are trained merely on the limited human-labeled data can improve themselves significantly.
We also empirically find that models trained with DPO loss can generate pseudo-labeled data with higher quality and diversity.
Our experiments demonstrate that the proposed method not only 
	outperforms existing methods with comparable model sizes on the GSM8K benchmark,
	but also achieves remarkable resource efficiency in terms of 
	both computational cost and the requirements of human-labeled data.

\section*{Limitations} \label{limit}

	\paragraph{Use of unlabeled data}
	Our method is built upon the classic self-training algorithm,
		which provides an effective semi-supervised learning framework
		capable of utilizing unlabeled data efficiently.
	However,
		this work doesn't explore the use of unlabeled data explicitly.
	Future research efforts can be made to explore
		how to collect high-quality unlabeled data for math word problem solving.
	In other words,
		we need to find an efficient method for collecting unlabeled data
		$\mathcal{U} = \{(x_i, a_i)\}_{i=1}^u$ that for each math question $x_i$, 
		there is a corresponding ground-truth answer $a_i$,
		ensuring the data's relevance and utility for enhancing model training.
		
	\paragraph{Generalization to other tasks}
	One of the limitations of this work is the narrow scope of our experiments, which were exclusively conducted on math reasoning tasks. 
	The primary reason for this limitation is the lack of appropriate training data for other reasoning tasks.
	As our method requires a set of training data with chain-of-thought labels,
		many existing reasoning tasks lack such annotations,
		making it challenging to extend our experiments beyond the current scope.
	Future research may focus on identifying and developing suitable datasets 
		for a wider range of reasoning tasks to fully evaluate the applicability and effectiveness of our method across different reasoning tasks.

\section*{Acknowledgements}

This work was done when Shichen Li was a visiting student at the StatNLP Research Group of SUTD.
We would like to thank the anonymous reviewers, our meta-reviewer, and senior area chairs for their constructive comments and support on this work. 
This research/project is supported by Ministry of Education, Singapore, under its Academic Research Fund (AcRF) Tier 2 Programme (MOE AcRF Tier 2 Award No: MOET2EP20122-0011), the National Research Foundation Singapore and DSO National Laboratories under the AI Singapore Program (AISG Award No: AISG2-RP-2020-016), and Ministry of Education, Singapore, under its Tier 3 Programme (The Award No.: MOET320200004). Any opinions, findings and conclusions or recommendations expressed in this material are those of the authors and do not reflect the views of the funding agencies.

\bibliography{custom}

\appendix
\section{Additional Implementation Details}\label{app:impl}
Our models are trained using the AdamW optimizer~\cite{loshchilov2017adamw}
    with a weight decay of 0.01 and gradient clipping of 1.0.
We employ a cosine learning rate schedule with warm-up.
During training, the maximum sequence lengths are set to 500 for T5 models and 640 for Llama models.
Both T5 and Llama models undergo DPO-ST for three iterations, using the same set of hyperparameters for each iteration as detailed in Table~\ref{tab:train_hp}.
For each DPO step, we sample 5 pseudo-labels per question from the SFT model to build the DPO training data, and set $\beta=0.1$ during DPO training.
In SFT steps, the number of model-generated solutions per question can be varied and controlled by the hyperparameter $K$.
When sampling pseudo-labels, we limit the maximum generated tokens to 300 and use a temperature of 0.7.

\begin{table}[h]
    \begin{center}
    \centering
    \resizebox{0.9\linewidth}{!}{
        \begin{tabular}{lcccc}
        \toprule
                    & \multicolumn{2}{c}{\bf{Flan-T5}}  & \multicolumn{2}{c}{\bf{LLaMA}} \\
				   \cmidrule(lr){2-3}                  \cmidrule(lr){4-5}
		\bf{Hyperparameters} & SFT & DPO & SFT & DPO \\
        \midrule
            Batch size       &   96      &   96         &    128    & 128    \\
            Epochs           &    8      &    -         &    2      & -       \\
		Max steps        &    -      &    150       &    -      & 100       \\
            Learning rate    &   3e-4    &   7e-7       &    2e-5   & 3e-7       \\
		Warm-up ratio    &   0.1     &   0.1        &    0.03   &    0.03      \\
        \bottomrule
        \end{tabular}
    }
    \caption{
        \label{tab:train_hp}
        Training details of SFT and DPO steps for Flan-T5 and Llama models.
    }
    \end{center}
\end{table}

\end{document}